\documentclass{article}
\usepackage{amsmath,graphicx,mlspconf}
\usepackage[numbers]{natbib}
\usepackage{microtype}
\usepackage{graphicx}
\usepackage{booktabs} 

\usepackage{nohyperref}
\newcommand\url[1]{#1}

\usepackage{amsmath}
\usepackage{amssymb}
\usepackage{amsthm}
\usepackage{graphicx}
\usepackage{subcaption}
\usepackage[inline]{enumitem}
\usepackage[utf8]{inputenc}
\DeclareUnicodeCharacter{00B1}{\pm}

\newtheorem{cor}{Corollary}
\newtheorem{prop}{Proposition}

\makeatother
\copyrightnotice{978-1-7281-6662-9/20/\$31.00 {\copyright}2020 IEEE}

\toappear{2020 IEEE International Workshop on Machine Learning for Signal Processing, Sept.\ 21--24, 2020, Espoo, Finland}

\title{Fixing Bias in Reconstruction-based Anomaly \\ Detection with Lipschitz Discriminators}
\name{
    Alexander Tong$^{\star}$%
    \qquad Guy Wolf$^{\dagger}$%
    \qquad Smita Krishnaswamy$^{\ddag \star}$\thanks{This research was partially funded by IVADO (l'institut de valorisation des donn\'{e}es) [\emph{G.W.}]; Chan-Zuckerberg Initiative grants 182702 \& CZF2019-002440 [\emph{S.K.}]; and NIH grants R01GM135929 \& R01GM130847 [\emph{G.W., S.K.}]. The content provided here is solely the responsibility of the authors and does not necessarily represent the official views of the funding agencies. Correspondence to: Smita Krishnaswamy \textless{} smita.krishnaswamy@yale.edu\textgreater{}}%
}
\address{%
    $^{\star}$ Yale University, Department of Computer Science; $^{\ddag}$ Department of Genetics, New Haven, CT, USA \\%
    $^{\dagger}$ Universit\'{e} de Montr\'{e}al, Dept. of Math. \& Stat.; Mila -- Quebec AI Institute, Montreal, QC, Canada\\%
}

\begin{document}

\maketitle

%

\begin{abstract}


Anomaly detection is of great interest in fields where abnormalities need to be identified and corrected (e.g., medicine and finance). Deep learning methods for this task often rely on autoencoder reconstruction error, sometimes in conjunction with other errors. We show that this approach exhibits intrinsic biases that lead to undesirable results. Reconstruction-based methods are sensitive to training-data outliers and simple-to-reconstruct points. Instead, we introduce a new unsupervised \textit{Lipschitz anomaly discriminator} that does not suffer from these biases. Our anomaly discriminator is trained, similar to the ones used in GANs, to detect the difference between the training data and corruptions of the training data. We show that this procedure successfully detects unseen anomalies with guarantees on those that have a certain Wasserstein distance from the data or corrupted training set. These additions allow us to show improved performance on MNIST, CIFAR10, and health record data.


\end{abstract}

\section{Introduction}

A common problem in real-world data analysis is to identify differences between samples in complex high-dimensional data with scarcity in labels or annotations. Here, we focus on the problem of \textit{unsupervised anomaly detection}, also known as outlier or novelty detection. We consider anomalous points as those that have a low likelihood of occurring in data generated from the nominal distribution, and are likely generated through some other (anomalous) process. This formulation suggests a density estimation solution. However, in high-dimensional datasets direct density estimation suffers from the well-known curse of dimensionality, requiring a sample size exponential in the dimension. This motivates the use of deep networks, which have proven effective in high dimensional spaces. 

Autoencoders are common in unsupervised anomaly detection with reconstruction error serving as a way to identify anomalies, as it gives worse reconstruction points that are far from the training set~\cite{chalapathy_deep_2019}. However, recent work studying overparameterized autoencoders suggests that MSE trained autoencoders tend to memorize their training set, essentially becoming one-nearest-neighbor density estimators~\cite{radhakrishnan_memorization_2019}. While they may converge to the true density with infinite samples~\cite{zhao_anomaly_2009}, this memorization can be detrimental for anomaly detection in realistic data. In particular, autoencoders are sensitive to outliers in the training data and biased towards easily reconstructed points, even if they are anomalous (e.g., in low-density regions of the convex hull of the data).

Some recent work (e.g., \cite{chalapathy_robust_2017, abati_latent_2019, sabokrou_adversarially_2018}) proposes the addition of loss terms that constrain the autoencoder latent space to better limit the capacity, while other works propose leveraging adversarial losses to compute reconstruction error in the latent space~\cite{schlegl_unsupervised_2017}. However, these methods all still fundamentally rely on a reconstruction penalty (with its inherent deficiencies) for quantifying abnormality of evaluated points compared to training data. 

We propose a new model that does not rely on reconstruction error, but rather on a discriminator network with a direct anomaly score. Our method explicitly encourages low anomaly scores over the training data, and high anomaly scores over a chosen corruption of the data. We show our method has theoretically well-defined behavior outside of the data, is less sensitive to outliers in the training data, and is more discriminative between points within the hull of the data. 

\begin{figure}[!b]
\centering
\begin{subfigure}{.33\linewidth}
  \centering
  \includegraphics[width=.9\linewidth]{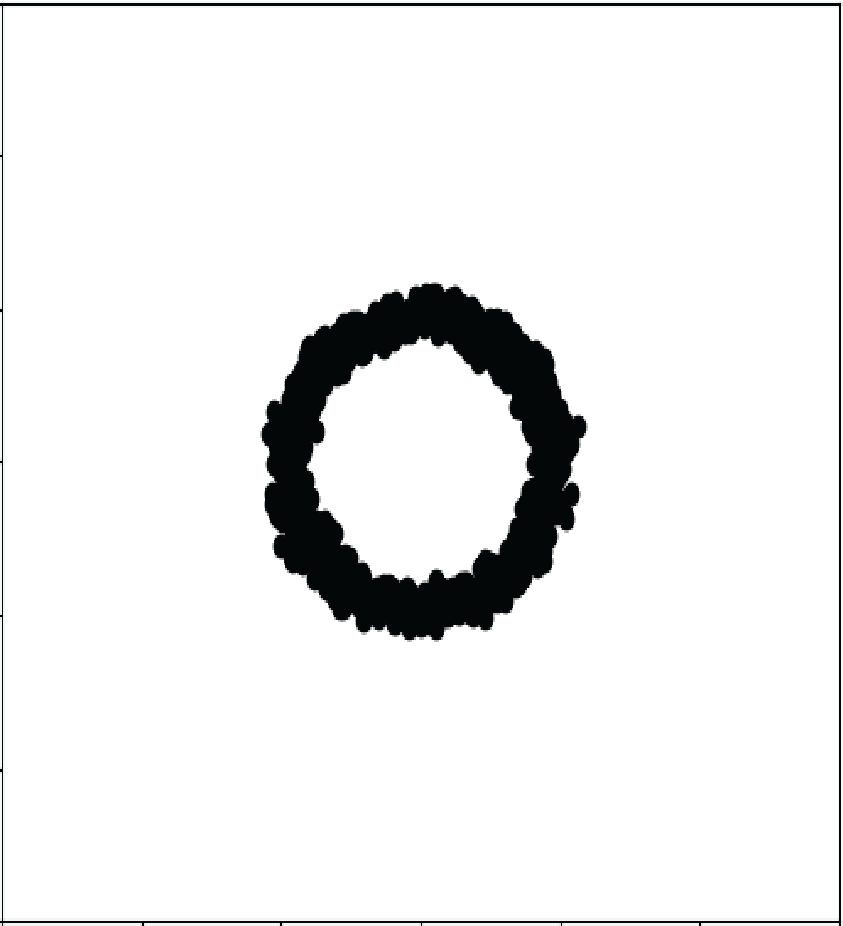}
  \caption{Input Data}
  \label{fig:circle:sfig1}
\end{subfigure}%
\begin{subfigure}{.33\linewidth}
  \centering
  \includegraphics[width=.9\linewidth]{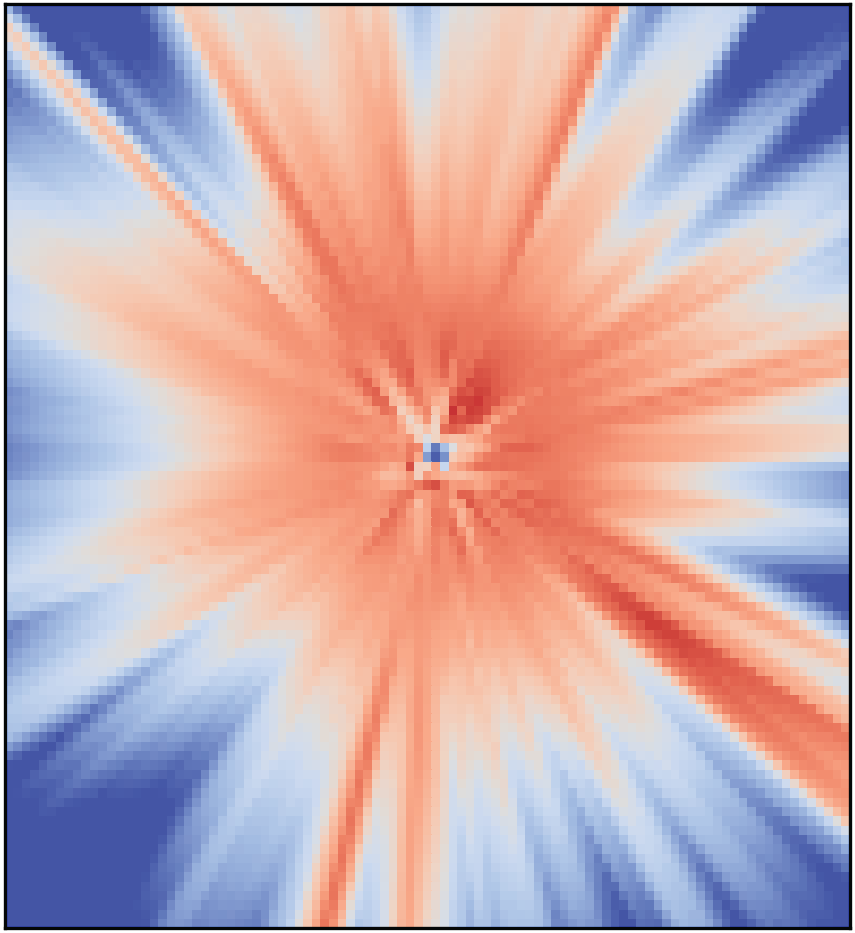}
  \caption{AE}
  \label{fig:circle:sfig2}
\end{subfigure}
\begin{subfigure}{.33\linewidth}
  \centering
  \includegraphics[width=.9\linewidth]{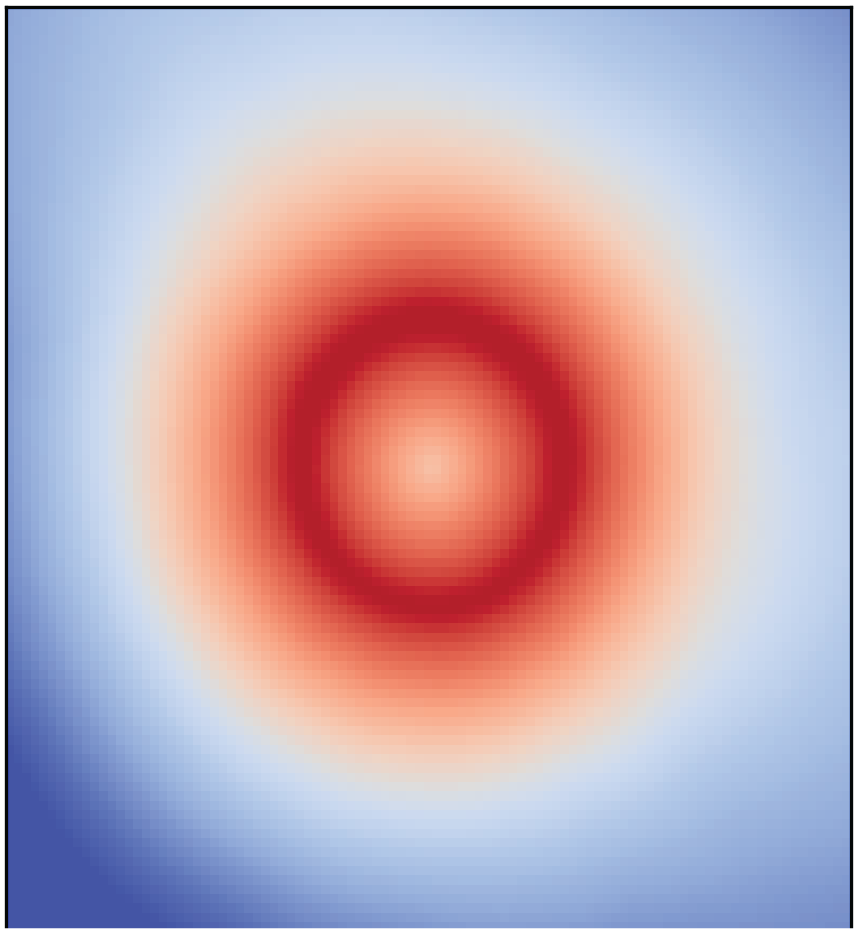}
  \caption{LAD}
  \label{fig:circle:sfig3}
\end{subfigure}
\vspace{-3mm}
\caption{(a) training data (b) autoencoder reconstruction score (c) LAD anomaly score (red = more normal, blue = more anomalous). Reconstruction is a poor proxy for data density and interpolates too well close to the data.}
\label{fig:circle}
\end{figure}



\section{Background} \label{sec:background}


Given samples from a nominal probability distribution $P$ over $\mathcal{X}$, the \textit{density level set} formulation can be seen as the following. Given some $\alpha > 0$ we wish to produce a decision function $c : \mathcal{X} \rightarrow \{0,1\}$ on the nominal probability $p(x)$, such that $c(x) = [[p(x) > \alpha]]$ where $\alpha$ is either predefined using some absolute density level or in relation to some quantile of the training data controlling the false positive error~\cite{zhao_anomaly_2009}. Existing methods use a corruption parameter $\gamma \in [0, 0.5)$ to set $\alpha$, based on scores on the training data.



Recently, there is renewed interest in the use of transportation metrics, such as the Wasserstein metric, to train neural networks. The Wasserstein GAN~\cite{arjovsky_wasserstein_2017} and later works \cite{gulrajani_improved_2017} demonstrated the advantages of training GANs with a Wasserstein based loss. The Wasserstein-1 metric, also known as Earth-mover distance, between two distributions $P$ and $Q$ defined over $\mathbb{R}^n$ with some distance metric $d$ is:
\vspace{-2mm}
\begin{equation}
    W(P,Q) = \inf_{\pi \in \Pi(P,Q)} \mathbb{E}_{(x,y) \sim \pi} \bigl [ d(x,y) \bigr ]
    \vspace{-2mm}
\end{equation}
This can be thought of as the minimum amount of work required to move one pile of dirt ($P$) to another ($Q$). However, the optimization over the infimum is intractable. By the Kantorovich-Rubinstein duality~\cite{villani_optimal_2009} for the Euclidean distance metric, i.e., $d(x,y) = \|x - y\|$,
\vspace{-2mm}
\begin{equation}
    W(P,Q) = \sup_{\|f\|_L \le 1} \mathbb{E}_{x \sim P} f(x) - \mathbb{E}_{x \sim Q} f(x)
    \vspace{-2mm}
\end{equation}
Here $f$ is often called the witness function, as it ``witnesses'' the difference between $P$ and $Q$ maximally. By constraining the network to be 1-Lipschitz we can, by gradient descent, approximate the Wasserstein distance between the two distributions for which we have samples.


\begin{figure}[!b]
    \begin{center}
    \includegraphics[width=1 \linewidth]{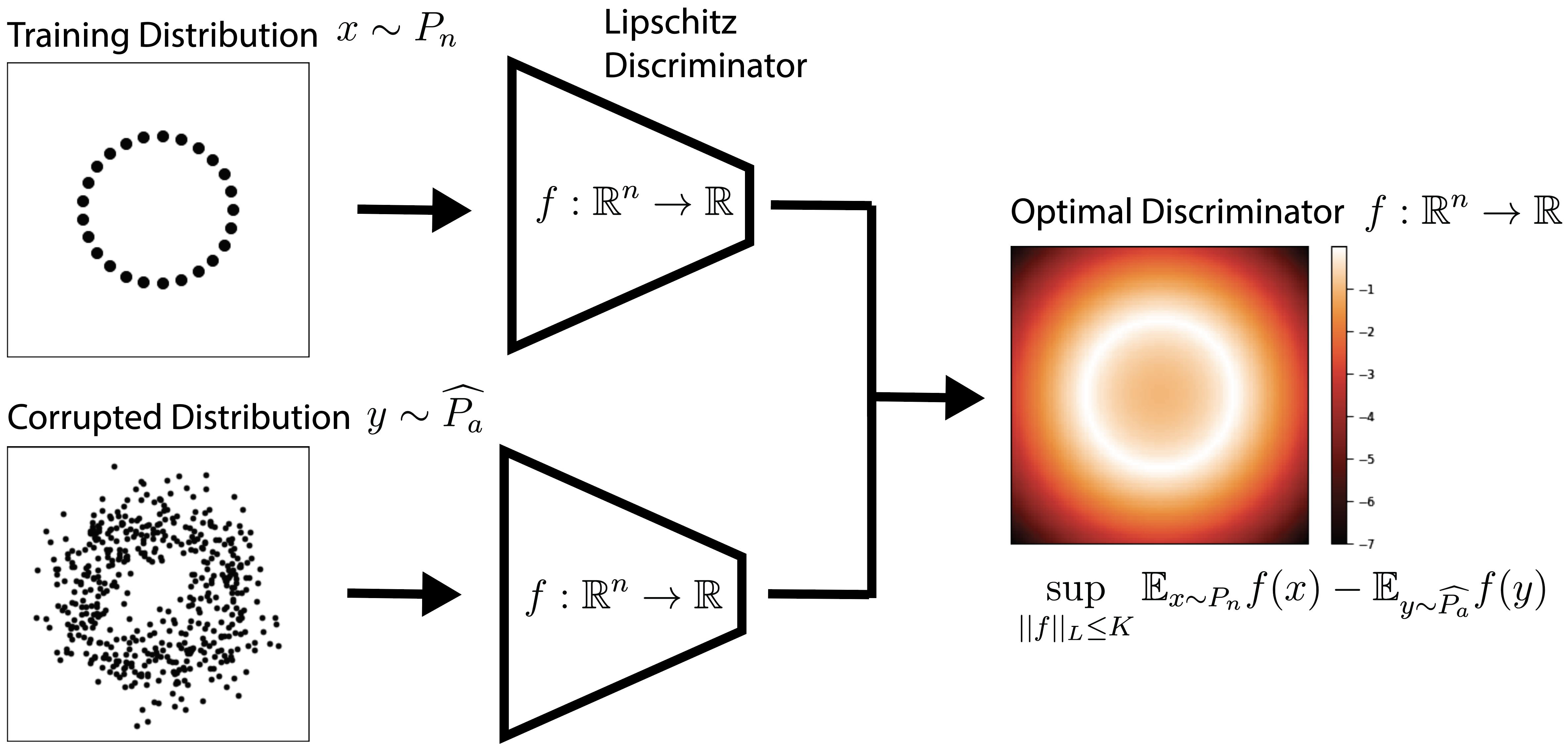}
    \end{center}
    \vspace{-3mm}
    \caption{LAD trains a Lipschitz neural network $f$ to discriminate between the data and a corrupted version of the data. Our trained network $f^*$ is then used to score anomalies. Darker = more anomalous.}
    \label{fig:schematic}
\end{figure}

\section{The Lipschitz Anomaly Discriminator}\label{sec:methods}

We propose to learn an anomaly scoring function $f$ as the output of a neural network we call \textit{Lipschitz Anomaly Discriminator} (LAD). Our neural network function $f$ is trained to maximally discriminate between nominal training data $P_n$, and a corrupted version of it $\widehat{P_a}$. Since we are tackling the unsupervised anomaly detection problem, we use a corrupted version of the training data as a substitute for the true (unknown) distribution of anomalous points $P_a$ (See Fig.~\ref{fig:schematic}). 

Rather than using an ordinary discriminator as in the standard generative adversarial network (GAN) framework, we use a Lipschitz constrained network as in \cite{arjovsky_wasserstein_2017} because this allows us to directly optimize for low normality scores on low density points on $P_n$ and high normality scores on high density points. Furthermore, a $K$-Lipschitz neural network optimized to discriminate between two distributions has a convenient formulation in terms of the Wasserstein distance and the Kantorovich-Rubenstein duality~\cite{villani_optimal_2009}. Note that we do not use the generator from \cite{arjovsky_wasserstein_2017}, only the discriminator. We use the training data, and generate a corrupted distribution by sampling from some other distribution $\widehat{P_a}$, meant to model the anomaly distribution as closely as possible. While we could use the samples of a generator to model $\widehat{P_a}$, we found more success using a simple corruption process. We use the gradient penalty formulation in WGAN-GP~\cite{gulrajani_improved_2017} to optimize our objective, given by:
\vspace{-2mm}
\begin{align}
    \label{eq:opt}
    L &= \mathbb{E}_{x \sim P_n} \bigl [f(x)\bigr] - \mathbb{E}_{x \sim \widehat{P_a}} \bigl [ f(x) \bigr] \\
      &+ \lambda \mathbb{E}_{x \sim P_x} \bigl [( \| \nabla_x f(x)\|_2 - 1)^2 \bigr] \, , \nonumber
    \vspace{-2mm}
\end{align}
where $P_x$ is obtained by sampling uniformly along straight lines between pairs of points sampled from $\widehat{P_a}$ and $P_n$. We use $\lambda=10$ as suggested in the original work.

\subsection{Estimating the Anomaly Distribution}
The choice of the anomaly distribution to train against is important and useful in building inductive bias into the model. Existing models implicitly build in an assumption on the anomaly distribution. For example, overparameterized autoencoders assume points are far from the span of the data~\cite{radhakrishnan_memorization_2019}.

We show the difference in the average score on the true anomalies and the average score on the true normal is bounded above by the Wasserstein distance between the true anomaly distribution and the estimate of the anomaly distribution (see Eq.~\ref{eq:stability}). Practically, this means that the better the estimate of the anomaly distribution, the better the performance of LAD. 

In our experiments, we choose Gaussian noise on the artificial cases, and to shuffle patches on images for MNIST and CIFAR10. In general, the closer the corruption distribution is to the true anomaly distribution, the better guarantees we can make. This is similar to models like denoising autoencoders or GANs where theoretically the noise should affect the performance, but in practice standard Gaussian noise performs well enough.

\subsection{Theoretical Properties}\label{sec:theory}

\subsubsection{Training Set Contamination} The standard anomaly detection task assumes access to training data sampled i.i.d. from some nominal distribution, which is unrealistic in a big data setting. Formally, we consider the problem where samples are drawn from a mixture of the nominal probability distribution $P_n$ and the anomalous distribution $P_a$. That is we are given $n$ samples $\{x_i\}_{i=1}^n$ drawn i.i.d.\ from the probability distribution $(1 - \gamma) P_n + \gamma P_a$, where $\gamma \in (0,1)$ and $\gamma \ll 1$ represents the anomaly contamination. We show that while existing deep anomaly detection methods perform well on clean training data (i.e., $\gamma=0$), they are very sensitive to even a small amount of contamination in the training set.


Suppose our training set is corrupted with a fraction $\gamma > 0$ of anomalous datapoints, then our method should still effectively distinguish between nominal and anomalous data. Because we are training the discriminator $f$ to maximize the difference between the nominal and corrupted points subject to a Lipschitz constraint, then the Kantorovich-Rubinstein duality holds. Implying that if our corruption process follows the distribution of anomalies (i.e., $\widehat{P_a} = P_a$) then the change in the difference between the nominal and anomalous points is bounded from below. This is summarized in the following proposition.
\begin{prop} 
\label{prop:robustness}
Let $f^{(A,B)}$ denote the optimal solution to $\max_{\|f\|_L \le 1} \mathbb{E}_{x \sim A} f(x) - \mathbb{E}_{x \sim B} f(x)$, and in particular let $f^* = f^{(P_n, P_a)}$ and $f^{**} = f^{((1 - \gamma) P_n + \gamma P_a, P_a)}$. Then, under the same conditions as Prop.~\ref{prop:threshold},
\begin{align*}
\bigl |\mathbb{E}_{x \sim P_n} [f^*(x) - f^{**}(x)] + \mathbb{E}_{x \sim P_a} [f^{**}(x) - f^{*}(x)] \bigr | \\
\le \frac{1}{1 - \gamma} W(P_n, (1-\gamma) P_n + \gamma P_a)
\end{align*}
\end{prop}

\begin{proof}
Let $\mathcal{A} = \bigl |\mathbb{E}_{x \sim P_n} [f^*(x) - f^{**}(x)] - \mathbb{E}_{x \sim P_a} [f^*(x) - f^{**}(x)] \bigr |$. Then, we can write 
\vspace{-2mm}
\begin{align*}
\mathcal{A} &= \bigl |\mathbb{E}_{x \sim P_n} [f^*(x) - f^{**}(x)] - \mathbb{E}_{x \sim P_a} [f^*(x) - f^{**}(x)] \bigr | \\
&= \bigl |\mathbb{E}_{x \sim (1-\gamma) P_n + \gamma P_a} [f^*(x) - f^{**}(x)] \\
&\quad- \mathbb{E}_{x \sim P_a} [f^*(x) - f^{**}(x)] + \gamma \mathcal{A} \bigr |,
\end{align*}
which forms a geometric series. Therefore, for $\gamma < 1$ we get
\begin{align*}
\mathcal{A} &= \frac{1}{1 - \gamma} \bigl |\mathbb{E}_{x \sim (1-\gamma) P_n + \gamma P_a} [f^*(x) - f^{**}(x)] \\
&\quad- \mathbb{E}_{x \sim P_a} [f^*(x) - f^{**}(x)] \bigr | .
\end{align*}

\noindent We now examine the $f^*$ and $f^{**}$ portions of $\mathcal{A}$ separately. By reorganizing terms, we can write $\mathcal{A} = \frac{1}{1 - \gamma} \bigl | \mathcal{A}_{f^*} + \mathcal{A}_{f^{**}} \bigr |$ with
\begin{align*}
\mathcal{A}_{f^*} &= \mathbb{E}_{x \sim (1-\gamma) P_n + \gamma P_a} [f^*(x)] - \mathbb{E}_{x \sim P_a} [f^*(x)] \\
&= (1-\gamma) \left( \mathbb{E}_{x \sim P_n} [f^*(x)] - \mathbb{E}_{x \sim P_a} [f^*(x)] \right) \\
&= (1-\gamma) W(P_n, P_a); \\
\mathcal{A}_{f^{**}} &= \mathbb{E}_{x \sim P_a} [f^{**}(x)] - \mathbb{E}_{x \sim (1-\gamma) P_n + \gamma P_a} [f^{**}(x)]\\
&= -W((1-\gamma) P_n + \gamma P_a, P_a).
\end{align*}
Combining terms and applying the triangle inequality,
\begin{align*}
&\mathcal{A} = \frac{1}{1 - \gamma} \bigl | \mathcal{A}_{f^*} + \mathcal{A}_{f^{**}} \bigr | \\
&= \frac{1}{1 - \gamma} \bigl | (1-\gamma) W(P_n, P_a) - W((1-\gamma) P_n + \gamma P_a, P_a) \bigr | \\
&\le \frac{1}{1 - \gamma} W(P_n, (1-\gamma) P_n + \gamma P_a) ,
\end{align*}
which proves the proposition.
\end{proof}
This result can be thought of bounding the difference of score on between normal points and anomalous points when the training set is corrupted by the addition of anomalies $0 \leq \gamma \leq 1$. When the training set is corrupted by a small amount, the anomaly scoring function does not change too much in expectation. Compared to a score based on reconstruction error, a single point can affect the scoring function an arbitrary large amount. For example, take an anomaly distribution that is a Dirac at one point which is easily reconstructable, e.g. the black image in MNIST. Then, corrupting the training data with a single anomalous example can take the MSE to zero for the entire anomaly distribution. This is an adversarial case, however, we find that this also occurs in practice (See Sec.~\ref{sec:exp:contamination}).

\subsubsection{Robustness to distant points} As is apparent in Fig.~\ref{fig:circle}, for reconstruction-based anomaly detection methods, we are not guaranteed to correctly predict anomalous points even for points that are very far from the support of the nominal data distribution. On the other hand, LAD does not suffer from this instability. We formalize this in our next proposition. Intuitively, points sufficiently far away from the data will have a higher anomaly score than any point in the data.

\begin{prop}
\label{prop:threshold}
Let $f^*$ be the optimal solution of \\
$\sup_{\|f\|_L \le 1} \left( \mathbb{E}_{x \sim P_n} [f(x)] - \mathbb{E}_{y \sim P_a} [f(y)] \right)$, 
and let $\pi$ be the optimal coupling between $P_n$ and $P_a$ defined as the minimizer of $W(P_n, P_a) = \inf_{\pi \in \Pi(P_n, P_a)} \mathbb{E}_{(x,y) \sim \pi} [ \|x - y\| ]$, where $\Pi(P_n, P_a)$ is the set of joint distributions whose marginals are $P_n$ and $P_a$, respectively. If $P_n$ has a compact support $\mathcal{S}_n$ and $P_a$ has compact support $\mathcal{S}_a$, then there exists $C > 0$ such that $f^*(y) \leq C - \inf_{x \in \mathcal{S}_n}\{\|x - y\|\}$ for $P_a$-almost every $y$.
\end{prop}

\begin{proof}
First, we recall from Theorem 5.10 (iii) of \cite{villani_optimal_2009} that since $P_n$ and $P_a$ have compact support $f^*$ exists, and from Theorem 5.10 (ii) of \cite{villani_optimal_2009} that $\pi$-almost surely, $f^*(x) - f^*(y) = \|x - y\|$, and therefore also $\pi$-a.s., $f^*(y) = f^*(x) - \|x - y\|$. Now, let 
\begin{equation}
A = \{y : \exists x \in \mathcal{S}_n \text{ s.t. } f^*(y) = f^*(x) - \|x - y\| \},
\end{equation}
and let $A^c$ be its complement. Then, clearly by definition, for any $(x,y) \in \mathcal{S}_n \times A^c$ we must have $f^*(y) \neq f^*(x) - \|x - y\|$, and thus $\pi(\mathcal{S}_n \times A^c) = 0$. Therefore, by the equality of marginals (and since $\mathcal{S}_n$ is the support of $P_n$, we have $P_a(A^c) = \pi(\mathcal{S}_n \times A^c) = 0$, which yields $P_a(A) = 1$. Thus, for $P_a$-almost every $y$ we can write $f^*(y) = f(x^\prime) - \|x^\prime - y\|$ for some $x^\prime \in \mathcal{S}_n$. Finally, since $\mathcal{S}_n$ is compact and $f^*$ is continuous, we can set $C = \max_{x \in f^*(x)} f^*(x)$ and clearly have both $f^*(x^\prime) \leq C$ and $\|x^\prime - y\| \geq \inf_{x \in \mathcal{S}_n}\{\|x - y\|\}$, which yields the result in the proposition.
\end{proof}

\begin{cor}
\label{corr:threshold}
Under the conditions of Prop.~\ref{prop:threshold}, there exists $R > 0$ such that for $P_a$-almost every $y$, if $\inf_{x \in \mathcal{S}_n} \|y - x\| > R$ then $f^*(y) < f^*(x)$ for $P_n$-almost every $x$. 
\end{cor}
\begin{proof}
Let $R$ be the diameter of $\mathcal{S}_n$ defined as $\sup_{x,y \in \mathcal{S}_n}\{\|x - y\|\}$, which must be finite since $\mathcal{S}_n$ is compact, and let $x_0 \in \mathcal{S}_n$ be a point where $f^*$ reaches its maximum value, which was chosen as $C$ in the proof of Prop.~\ref{prop:threshold}. Then, since $f^*$ is a Lipschitz-continuous function, then for every $x \in \mathcal{S}_n$ we have $0 \leq C - f(x) \leq \|x - x_0\| \leq R$, thus, $f(x) \geq C - R$, and together with Prop.~\ref{prop:threshold} we get the result of the corollary.
\end{proof}

For a single training example, this can be understood as a peak at the training example, with $f^*$ decaying with slope 1 in every direction from this point. If we zoom out enough, every set of bounded support is similar to this single training example.

As a final note, we would like to bound the difference between the optimal discriminator function and the one learned by our estimate $\widehat{P_a}$ of the unknown true $P_a$. We note that this can be achieved directly from the triangle inequality over the Wasserstein distance. Indeed, since both $W(P_n, P_a) \le W(P_a, \widehat{P_a}) + W(\widehat{P_a},P_n)$ and $W(P_n, \widehat{P_a}) \le W(P_n, P_a) + W(P_a, \widehat{P_a})$, then 
\begin{align}
    \label{eq:stability}
    \bigl |W(P_n, P_a) - W(P_n, \widehat{P_a}) \bigr 
    |
    &\le W(P_a, \widehat{P_a}) \, .
\end{align}
Our model is bounded close to the optimal discriminator depending on the choice of anomaly and corruption distributions.

\section{Experimental Evaluation}\label{sec:eval}

We compare LAD against a selection of reconstruction-based methods (CAE, DCAE, RCAE, AND)~\cite{chalapathy_robust_2017, hawkins_outlier_2002, vincent_stacked_2010-1}, reconstruction-based methods with adversarial additions (ALOCC, AnoGAN)~\cite{schlegl_unsupervised_2017, sabokrou_adversarially_2018}, support vector methods (OCSVM, DSVDD)~\cite{scholkopf_estimating_2001, ruff_deep_2018}, and more traditional methods (IF, LOF)~\cite{liu_isolation-based_2012, breunig_lof_2000}. AND and AnoGAN models were not run for this paper so results are not available for training set contaminated instances (See Table~\ref{tab:mnist-corrupt}). Our experiments are intended to compare LAD with existing reconstruction-based methods as well as traditional anomaly detection methods.

We  show how reconstruction-based methods are sensitive to training set contamination (Sec.~\ref{sec:exp:contamination}) and perform poorly on interpolated points (Sec.~\ref{sec:exp:robust}). Finally, we show how combined with a standard convolutional autoencoder, LAD can achieve excellent performance on MNIST and CIFAR10.


To evaluate our method, we apply our model to three datasets, MNIST, CIFAR10, and the veterans aging cohort study (VACS), an electronic health record dataset with lab values for 1.3 million visits of HIV-positive veterans. For an evaluation metric, we use the area under the curve (AUC) of the receiver operator characteristics (ROC) curve. Using the AUC allows for fair comparison between scoring functions disregarding thresholding. All error bars are 95\% confidence intervals. All quantitative experiments were run over 3 initializations. All networks have depth 3 with LeakyReLU activations and widths are chosen to match the number of parameters within 1\%. Convolutional layers were used for MNIST and CIFAR10 and dense layers were used for VACS.

\subsection{Training Set Contamination}\label{sec:exp:contamination} 

To demonstrate empirically the problem with contamination, we show what happens on a standard MNIST anomaly detection task as we add some corruption to the training set. Previous work considers the mean AUC over classes trained on each one of the 10 digits with pure training sets. We consider the same with one additional variable, training set contamination $\pi \in [0, 0.10]$, where up to $10\%$ of the training set are other digits. To contaminate the training data for a given digit, we add random samples from the training data of the other 9 digits to the training data until we reach the correct fraction of training set contamination. To create the corrupted training distribution $\widehat{P_a}$, we take $4 \times 4$ patches of the image and shuffle these in random order. To evaluate each model, we use AUC over the entire test set containing 10,000 images, 1,000 of which are the nominal class. Examining the results in Table~\ref{tab:mnist-corrupt}, most methods perform quite well on a training set that only contains nominal data, but performance decays rapidly with increased training set corruption. LAD performs competitively on the standard task, but importantly, as the contamination increases above $1\%$, LAD continues to perform well, showing its robustness to training set contamination. This is expected in light of Prop.~\ref{prop:robustness}. When combined with a convolutional autoencoder as described in Sec.~\ref{sec:methods}, LAD+CAE outperforms existing methods on this task.

\begin{table}
\centering
\scalebox{0.8}{
\begin{tabular}{lrrrr|r}
\toprule
Train Corrupt. &   0.00 &   0.01 &   0.05 &    0.10 & Black \\
\midrule

ALOCC~\cite{sabokrou_adversarially_2018}        &  0.694 &  0.511 & 0.539 &  0.509 &  0.168 \\
AND~\cite{abati_latent_2019}                    &  0.975 &  -     & -     &  -     &  - \\
AnoGAN~\cite{schlegl_unsupervised_2017}                 & 0.913 &  -     & -     &  -     &  - \\
CAE~\cite{hawkins_outlier_2002}                 &  0.965 &  0.925 & 0.868 &  0.832 &  0.067 \\
DCAE~\cite{vincent_stacked_2010-1}              &  0.967 &  0.925 & 0.865 &  0.829 &  0.059 \\
DSVDD~\cite{ruff_deep_2018}                     &  0.748 &  0.788 & 0.718 &  0.696 &  0.571 \\
IF~\cite{liu_isolation-based_2012}              &  0.853 &  0.853 & 0.837 &  0.822 &  0.312 \\
LOF~\cite{breunig_lof_2000}                     &  0.973 &  0.958 & 0.789 &  0.709 &  0.695 \\
OCSVM~\cite{scholkopf_estimating_2001}          &  0.954 &  0.895 & 0.828 &  0.794 &  0.677 \\
RCAE~\cite{chalapathy_robust_2017}              &  0.957 &  0.934 & 0.870 &  0.832 &  0.049 \\
LAD (ours)                                      &  0.940 &  0.937 & 0.923 &  0.901 & \textbf{1.000} \\
LAD + CAE (ours)                                & \textbf{0.981} & \textbf{0.965} & \textbf{0.936} & \textbf{0.912} & \textbf{1.000} \\
\bottomrule
\end{tabular}
}
\caption{Mean AUC over digits over 3 seeds for 0\%-10\% training set corruption on MNIST. Methods not re-implemented are marked with - and provided for context. Last column shows rank of all black image in nominal test set. Higher is better.}
\vspace{-3mm}
\label{tab:mnist-corrupt}
\end{table}

\begin{table}[ht]
\centering
\scalebox{0.8}{
\begin{tabular}{lrrrr|r}
\toprule
Class &  plane &  car &    bird &         dog &            mean \\
\midrule
ALOCC~\cite{sabokrou_adversarially_2018}    &  0.421 &  0.439 &  0.530 &  0.473 &  0.463 \\
AND~\cite{abati_latent_2019}                &  0.717 &  0.494 &  0.662 &  0.504 &  0.617 \\
AnoGAN~\cite{schlegl_unsupervised_2017}     &  0.671 &  0.547 &  0.529 &  \textbf{0.603} &  0.618 \\
CAE~\cite{hawkins_outlier_2002}             &  0.683 &  0.454 &  0.677 &  0.525 &  0.604 \\
DCAE~\cite{vincent_stacked_2010-1}          &  0.689 &  0.447 &  \textbf{0.679} &  0.526 &  0.605 \\
DSVDD~\cite{ruff_deep_2018}                 &  0.518 &  0.656 &  0.528 &  0.568 &  0.571 \\
IF~\cite{liu_isolation-based_2012}          &  0.670 &  0.442 &  0.645 &  0.516 &  0.599 \\
LOF~\cite{breunig_lof_2000}                 &  0.661 &  0.440 &  0.649 &  0.511 &  0.575 \\
OCSVM~\cite{chen_one-class_2001}            &  0.684 &  0.456 &  0.674 &  0.502 &  0.590 \\
RCAE~\cite{chalapathy_robust_2017}          &  0.675 &  0.429 &  0.669 &  0.531 &  0.592 \\
LAD (ours)                                  &  0.597 &  \textbf{0.663} &  0.411 &  0.561 &  0.565 \\
LAD + CAE (ours)                            &  \textbf{0.723} &  0.497 &  0.652 &  0.544 &  \textbf{0.635} \\

\bottomrule
\end{tabular}
}
\caption{AUC on CIFAR10 for representative classes over 3 seeds with no corruption.}
\vspace{-3mm}
\label{tab:cifar-corrupt}
\end{table}


\begin{figure*}[!t]
\centering
\begin{subfigure}{.32\linewidth}
  \centering
  \includegraphics[width=1\linewidth]{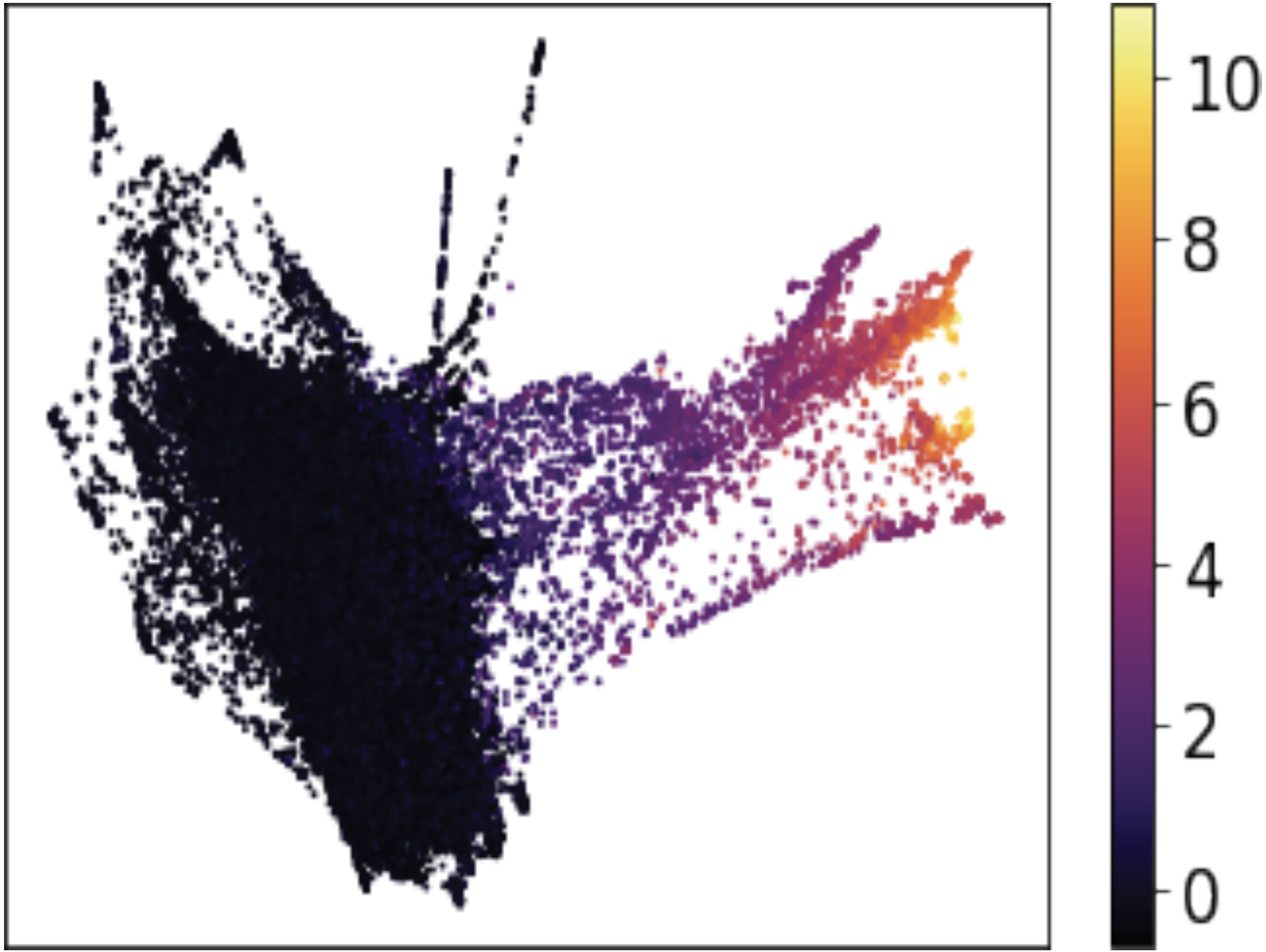}
  \caption{}
  \label{fig:vacs:sfig1}
\end{subfigure}%
\begin{subfigure}{.42\linewidth}
  \centering
  \includegraphics[width=1\linewidth]{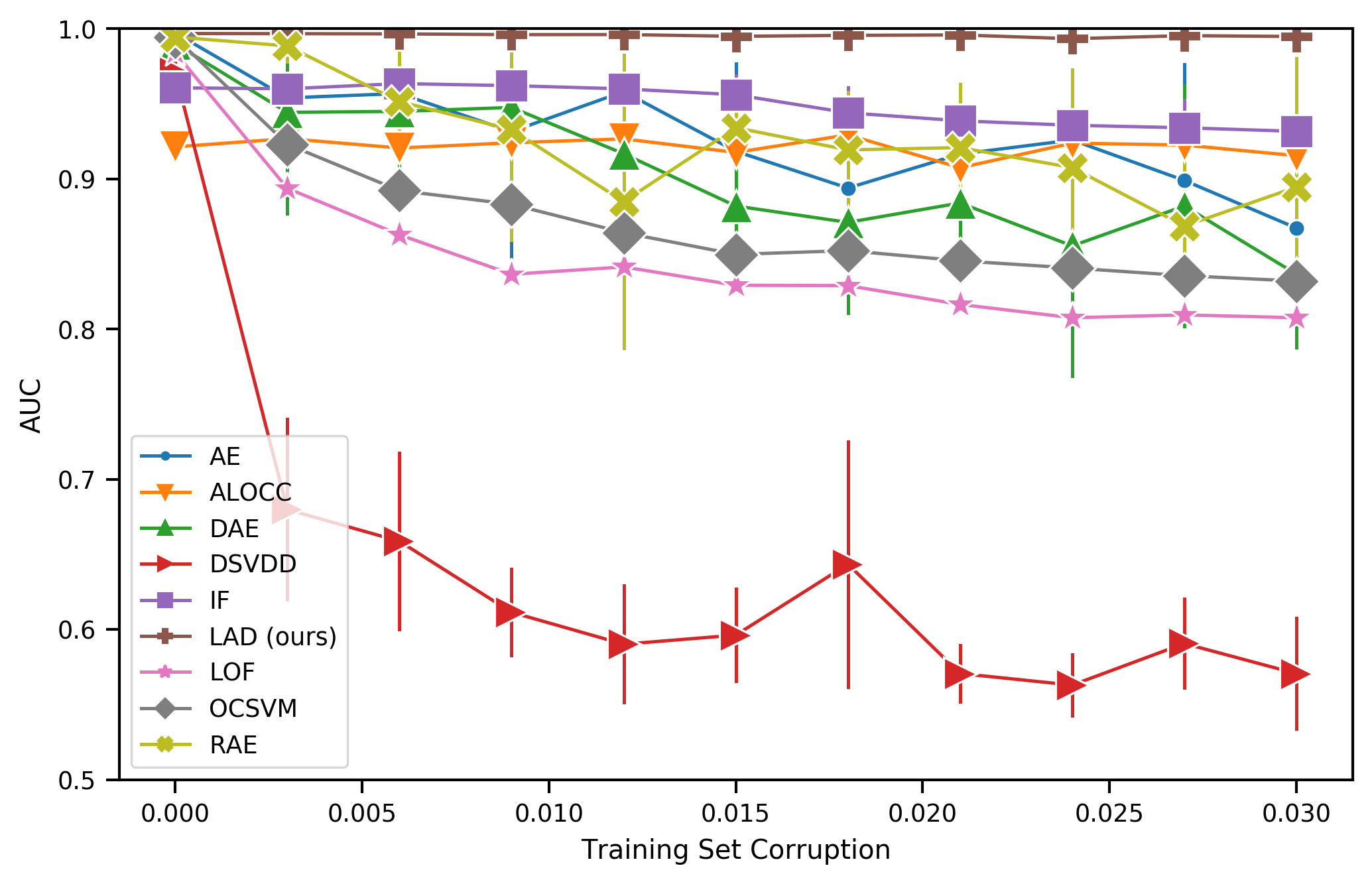}
  \vspace{-7mm}
  \caption{}
  \label{fig:vacs:sfig2}
\end{subfigure}
\vspace{-2mm}
    \caption{
    (a) Shows input data embedding with PHATE~\cite{moon_visualizing_2019} colored by Creatinine in mg/dL. (b) 
    Shows the AUC of various anomaly detection models over levels of training set corruption on the VACS data.}
    \label{fig:vacs}
\vspace{-3mm}
\end{figure*}

To establish usability in medical settings, we test our model on health records (see Fig.~\ref{fig:vacs}) that represent a large source of data that is error prone and difficult to cleanse of anomalies for training purposes. Here, we use the veterans aging cohort study (VACS) dataset\footnote{\url{https://medicine.yale.edu/intmed/vacs/}} containing 1.3 million clinic visits by over 28,000 HIV-positive veterans. Ten HIV relevant lab values were chosen and standardized (zero mean, unit std). To establish a set of known nominal vs.\ anomalous points we use clinic visits with a Creatinine lab value $>$ 2 std away from the mean (in this case $>$ 4.30 mg/dL) as anomalous. This threshold is well above the normal patient reference range \cite{inker_assessment_2018}, indicating high risk for renal disease. We split our data into 80\% training and validation set, and 20\% test set. To vary the training set contamination, we add a percentage of the high Creatinine values to the nominal training set. Since these patients are all in a similar state, even adding 0.3\% outliers to the training set drastically decreases performance. Fig.~\ref{fig:vacs} shows the AUC performance of various deep models on this task. Adding even a small amount of high Creatinine lab values encourages an autoencoder to represent them, and thereby all other anomalous ones, reducing performance for reconstruction error based models. Our model robustly detects high Creatinine anomalies, up to 3\% training set contamination. In medical applications where data cleaning is difficult, and outliers are often concentrated around specific values, LAD may be a better model choice.

\begin{figure}[!b]
\vspace{-4mm}
\centering
\begin{subfigure}{.35\linewidth}
  \centering
  \includegraphics[width=\linewidth]{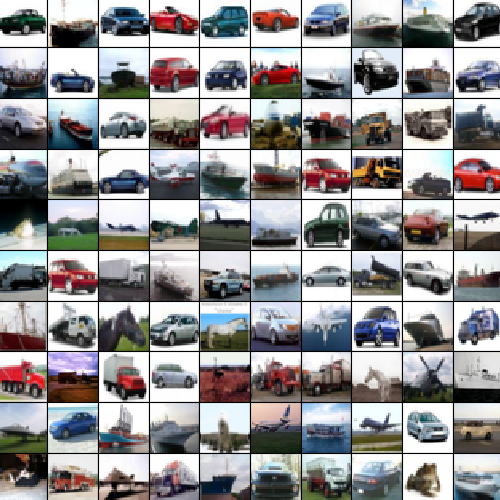}
  \caption{}
  \label{fig:cifar:sfig1}
\end{subfigure}%
\begin{subfigure}{.4\linewidth}
  \centering
  \includegraphics[width=.9\linewidth]{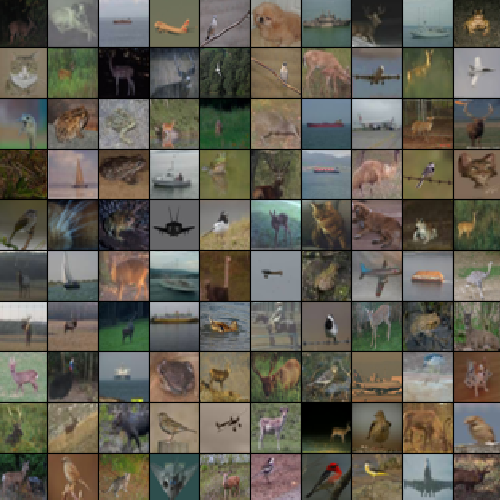}
   \caption{}
  \label{fig:cifar:sfig2}
\end{subfigure}
\begin{subfigure}{.165\linewidth}
  \centering
  \includegraphics[width=.9\linewidth]{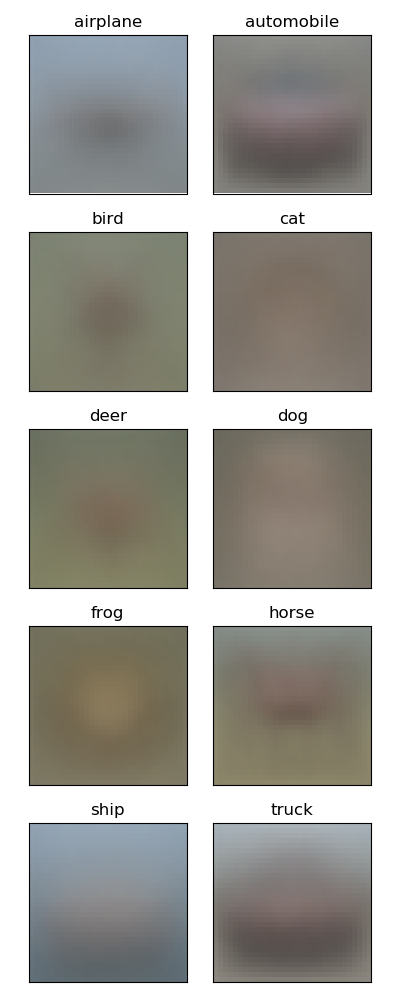}
   \caption{}
  \label{fig:cifar:sfig3}
\end{subfigure}
\vspace{-3mm}
\caption{Top 100 nominal images in test set of LAD (a) and DCAE (b) trained on the automobile class. (c) Mean over examples of pixel values for each class. Many car images have white background and/or bright colors that are far from the mean image. LAD does better at modeling such images.}
\label{fig:cifar}
\end{figure}

\subsection{Bias towards interpolated points}\label{sec:exp:robust}

Further experiments to study the interpolation bias are shown on MNIST with an all-black image, which is close to the mean MNIST image. Anomaly scores of the black image are compared to anomaly scores of nominal test images by computing the rank of its score within the nominal test data. We train the model on all `0's in the training set, and for testing we compute the black-image score relative to that of the 1,000 test `0's, with higher rank being more anomalous. We would expect this image to be more anomalous than any of the images in the test set (i.e., have a rank of 1). Table~\ref{tab:mnist-corrupt} shows that models based on reconstruction error consider the all black image as less anomalous than many of the test digits. LAD consistently ranks the black image as most anomalous.

We perform the same training set corruption experiment on CIFAR10 to show performance on a more complicated dataset (see Table~\ref{tab:cifar-corrupt}). We find there is a large variability in performance of each model between classes. In four of the classes, LAD outperforms all other models. Most methods perform poorly on the automobile and truck classes, which is consistent with~\cite{abati_latent_2019, ruff_deep_2018}. Additionally, reconstruction-based methods give low anomaly scores to test points that are close to the mean training image, as it is relatively easy to reconstruct. This is a poor inductive bias in cases where the nominal distribution has low density near the mean. In images of animals most of them are fairly close to the mean. However, in objects with the potential for bright colors (e.g., red cars on white backgrounds or black planes on white sky) these nominal images are very far from the mean training image for the class. This is shown in Fig.~\ref{fig:cifar} where most nominal points predicted by a DCAE are relatively near to the mean image. In contrast, LAD gives more varied top nominal images often mistaking trucks for cars but not animals for cars. 



\section{Conclusion} \label{sec:conclusion}

In this work, we introduce the first Wasserstein-distance based deep anomaly detection framework. We show the advantage of our discriminator-based framework, which learns to reason about the probability density of the nominal data and produces an anomaly score based on distribution distances, as opposed to other deep methods that rely on reconstruction-error criteria. We show reconstruction methods may learn to reconstruct ``averaged images'' that are within the convex hull of the data. In contrast, LAD provides guarantees on points with low support in the training data. Furthermore, we showed that LAD significantly outperforms existing methods on slightly corrupted training data, which is more realistic than assuming anomaly-free training data in large datasets. When combined with a standard autoencoder we find LAD outperforms existing methods on MNIST, CIFAR10, and a health record dataset displaying the effectiveness of our method.




\bibliographystyle{IEEEbib}
\bibliography{MLSP}

\begin{thebibliography}{10}

\bibitem{chalapathy_deep_2019}
R.~Chalapathy and S.~Chawla,
\newblock ``Deep learning for anomaly detection,''
\newblock arXiv:1901.03407, 2019.

\bibitem{radhakrishnan_memorization_2019}
A.~Radhakrishnan, K.~Yang, M.~Belkin, and C.~Uhler,
\newblock ``Memorization in overparameterized autoencoders,''
\newblock arXiv:1810.10333, 2019.

\bibitem{zhao_anomaly_2009}
M.~Zhao and V.~Saligrama,
\newblock ``Anomaly detection with score functions based on nearest neighbor
  graphs,''
\newblock in {\em NeurIPS}, 2009.

\bibitem{chalapathy_robust_2017}
R.~Chalapathy, A.K. Menon, and S.~Chawla,
\newblock ``Robust, deep and inductive anomaly detection,''
\newblock in {\em ECML}, 2017.

\bibitem{abati_latent_2019}
D.~Abati, A.~Porrello, S.~Calderara, and R.~Cucchiara,
\newblock ``Latent space autoregression for novelty detection,''
\newblock in {\em CVPR}, 2019.

\bibitem{sabokrou_adversarially_2018}
M.~Sabokrou, M.~Khalooei, M.~Fathy, and E.~Adeli,
\newblock ``Adversarially learned one-class classifier for novelty detection,''
\newblock in {\em CVPR}, 2018.

\bibitem{schlegl_unsupervised_2017}
T.~Schlegl, P.~Seeb{\"o}ck, S.M. Waldstein, U.~{Schmidt-Erfurth}, and G.~Langs,
\newblock ``Unsupervised anomaly detection with generative adversarial networks
  to guide marker discovery,''
\newblock in {\em IPML}, 2017.

\bibitem{arjovsky_wasserstein_2017}
M.~Arjovsky, S.~Chintala, and L.~Bottou,
\newblock ``Wasserstein {GAN},''
\newblock in {\em ICML}, 2017.

\bibitem{gulrajani_improved_2017}
I.~Gulrajani, F.~Ahmed, M.~Arjovsky, V.~Dumoulin, and A.~Courville,
\newblock ``Improved training of {Wasserstein} {GANs},''
\newblock in {\em NeurIPS}, 2017.

\bibitem{villani_optimal_2009}
C.~Villani,
\newblock {\em Optimal Transport: Old and New},
\newblock Springer, Berlin, 2009.

\bibitem{hawkins_outlier_2002}
S.~Hawkins, H.~He, G.~Williams, and R.~Baxter,
\newblock ``Outlier detection using replicator neural networks,''
\newblock in {\em DaWaK}, 2002.

\bibitem{vincent_stacked_2010-1}
P.~Vincent, H.~Larochelle, I.~Lajoie, Y.~Bengio, and {P.-A.} Manzagol,
\newblock ``Stacked denoising autoencoders: Learning useful representations in
  a deep network with a local denoising criterion,''
\newblock {\em JMLR}, 2010.

\bibitem{scholkopf_estimating_2001}
B.~Sch{\"o}lkopf, J.C. Platt, J.~{Shawe-Taylor}, A.J. Smola, and R.C.
  Williamson,
\newblock ``Estimating the support of a high-dimensional distribution,''
\newblock {\em Neural Computation}, 2001.

\bibitem{ruff_deep_2018}
L.~Ruff, R.A. Vandermeulen, N.~G{\"o}rnitz, L.~Deecke, S.A. Siddiqui,
  A.~Binder, E.~M{\"u}ller, and M.~Kloft,
\newblock ``Deep one-class classification,''
\newblock in {\em ICML}, 2018.

\bibitem{liu_isolation-based_2012}
F.T. Liu, K.M. Ting, and {Z.-H.} Zhou,
\newblock ``Isolation-based anomaly detection,''
\newblock {\em KDD}, 2012.

\bibitem{breunig_lof_2000}
M.M. Breunig, {H.-P.} Kriegel, R.T Ng, and J.~Sander,
\newblock ``{LOF}: Identifying density-based local outliers,''
\newblock in {\em ACM SIGMOD}, 2000.

\bibitem{chen_one-class_2001}
Y.~Chen, X.~Zhou, and T.~Huang,
\newblock ``One-class {SVM} for learning in image retrieval,''
\newblock in {\em ICIP}, 2001, pp. 34--37.

\bibitem{moon_visualizing_2019}
K.R. Moon, D.~van Dijk, Z.~Wang, S.~Gigante, D.B. Burkhardt, W.S. Chen, K.~Yim,
  A.~van~den Elzen, M.J. Hirn, R.R. Coifman, N.B. Ivanova, G.~Wolf, and
  S.~Krishnaswamy,
\newblock ``Visualizing structure and transitions in high-dimensional
  biological data,''
\newblock {\em Nature Biotechnology}, 2019.

\bibitem{inker_assessment_2018}
L.A. Inker and R.D. Perrone,
\newblock {\em Assessment of Kidney Function}, 2018,
\newblock \url{www.uptodate.com/contents/assessment-of-kidney-function}.

\end{thebibliography}

\end{document}



\author{
  Alexander Tong\thanks{Yale University, Depts. of Genetics \& Comp. Sci.}
  \and Guy Wolf\thanks{Universit\'{e} de Montr\'{e}al, Dept. of Math. \& Stat. ; Mila}$~^{,}$\footnotemark[3]
  \and Smita Krishnaswamy\footnotemark[1]$~^{,}$\thanks{Equal contrib.; Corr. auth.: \texttt{smita.krishnaswamy@yale.edu}}
}
\date{}

\maketitle


\fancyfoot[R]{\scriptsize{Copyright \textcopyright\ 20XX by SIAM\\
Unauthorized reproduction of this article is prohibited}}




\begin{appendix}
\beginsupplement
\input{appendix/proofs.tex}

\input{appendix/table.tex}
\begin{table}[ht]
\centering
\scalebox{0.8}{
\begin{tabular}{lrrrr|r}
\toprule
Class &  plane &  car &    bird &         dog &            mean \\
\midrule
ALOCC~\cite{sabokrou_adversarially_2018}    &  0.421 &  0.439 &  0.530 &  0.473 &  0.463 \\
AND~\cite{abati_latent_2019}                &  0.717 &  0.494 &  0.662 &  0.504 &  0.617 \\
AnoGAN~\cite{schlegl_unsupervised_2017}     &  0.671 &  0.547 &  0.529 &  \textbf{0.603} &  0.618 \\
CAE~\cite{hawkins_outlier_2002}             &  0.683 &  0.454 &  0.677 &  0.525 &  0.604 \\
DCAE~\cite{vincent_stacked_2010-1}          &  0.689 &  0.447 &  \textbf{0.679} &  0.526 &  0.605 \\
DSVDD~\cite{ruff_deep_2018}                 &  0.518 &  0.656 &  0.528 &  0.568 &  0.571 \\
IF~\cite{liu_isolation-based_2012}          &  0.670 &  0.442 &  0.645 &  0.516 &  0.599 \\
LOF~\cite{breunig_lof_2000}                 &  0.661 &  0.440 &  0.649 &  0.511 &  0.575 \\
OCSVM~\cite{chen_one-class_2001}            &  0.684 &  0.456 &  0.674 &  0.502 &  0.590 \\
RCAE~\cite{chalapathy_robust_2017}          &  0.675 &  0.429 &  0.669 &  0.531 &  0.592 \\
LAD (ours)                                  &  0.597 &  \textbf{0.663} &  0.411 &  0.561 &  0.565 \\
LAD + CAE (ours)                            &  \textbf{0.723} &  0.497 &  0.652 &  0.544 &  \textbf{0.635} \\

\bottomrule
\end{tabular}
}
\caption{AUC on CIFAR10 for representative classes over 3 seeds with no corruption.}
\vspace{-3mm}
\label{tab:cifar-corrupt}
\end{table}



\input{appendix/generator_investigation.tex}
\input{appendix/architecture.tex}
\end{appendix}

\bibliography{auto_DO_NOT_MODIFY}